\documentclass{article} %
\usepackage{iclr2020_conference,times}

\usepackage{amsmath,amsfonts,bm}

\def\eqref#1{equation~\ref{#1}}

\def\1{\bm{1}}

\def\rva{{\mathbf{a}}}

\def\rvx{{\mathbf{x}}}

\DeclareMathAlphabet{\mathsfit}{\encodingdefault}{\sfdefault}{m}{sl}
\SetMathAlphabet{\mathsfit}{bold}{\encodingdefault}{\sfdefault}{bx}{n}

\usepackage{hyperref}
\usepackage{url}
\usepackage{graphicx}       %
\usepackage{enumitem}

\newlist{compactitem}{itemize}{3} %
\setlist[compactitem]{label=\textbullet, leftmargin=1em, nosep}

\newcommand{\xhdr}[1]{\vspace{1mm}\noindent{{\bf #1.}}}

\newcommand{\regact}{\grave{a}}
\newcommand{\cact}{\acute{a}}
\newcommand{\cactset}{\check{\rva}}
\newcommand{\utt}{\rvx}
\newcommand{\indutt}{I\left(\utt\right)}
\newcommand{\intexc}{T^{\left(0:L-1\right)}}

\title{Ecological Semantics: Programming Environments for Situated Language Understanding}

\author{Ronen Tamari \& Dafna Shahaf \\
The Hebrew University of Jerusalem\\
\texttt{\{ronent,dshahaf\}@cs.huji.ac.il} \\
\And
Gabriel Stanovsky \& Reut Tsarfaty \\
Allen Institute for Artificial Intelligence \\
\texttt{\{gabis,reutt\}@allenai.org} \\
}

\iclrfinalcopy %
\begin{document}

\maketitle

\begin{abstract}
Large-scale natural language understanding (NLU) systems have made impressive progress: they can be applied flexibly across a variety of tasks, and employ minimal structural assumptions. However, extensive empirical research has shown this to be a double-edged sword, coming at the cost of shallow understanding: inferior generalization, grounding and explainability. Grounded language learning approaches offer the promise of deeper understanding by situating learning in richer, more structured training environments, but are limited in scale to relatively narrow, predefined domains. How might we enjoy the best of both worlds: grounded, general NLU? Following extensive contemporary cognitive science, we propose treating environments as ``first-class citizens'' in semantic representations, worthy of research and development in their own right. Importantly, models should also be partners in the creation and configuration of environments, rather than just actors within them, as in existing approaches. To do so, we argue that models must begin to understand and program in the language of affordances (which define possible actions in a given situation) both for online, situated discourse comprehension, as well as large-scale, offline common-sense knowledge mining. To this end we propose an environment-oriented ecological semantics, outlining theoretical and practical approaches towards implementation.  We further provide actual demonstrations building upon interactive fiction programming languages.

\end{abstract}

\section{Introduction}
 \begin{quotation}
\noindent ``Ask not what's inside your head, but what your head's inside of.'' \citep{mace1977james}
 \end{quotation}

Recovery of meaning is at the heart of the endeavor to build better natural language understanding (NLU) systems. Semantics researchers study meaning representation, and in particular the relations between language and cognitive representations~\citep{gardenfors2014geometry}.

A recurring point of contention in semantics research~\citep{fodor1988connectionism,Mahon2008} concerns the degree to which knowledge representation and language comprehension involve a \textit{symbolic} internal language of thought (LoT)~\citep{fodor1975language} or are \textit{embodied}; i.e., grounded in the brain's systems for action and perception~\citep{Feldman2004,Barsalou2007}. 

Current deep-learning methods for large-scale NLU, such as BERT~\citep{devlin2018bert}, incorporate minimal cognitive biases and assume primarily distributional semantics~\citep{firth1957synopsis}. Extensive empirical research shows this to be a double-edged sword: while affording widespread applicability to a variety of tasks, such methods are limited by impoverished training environments (static datasets, narrow contextual prediction, etc.) and struggle in settings requiring deeper understanding, such as systematic generalization~\citep{lake2019human,McCoy2019BERTsOA}, common-sense~\citep{forbes2019neural} and explainability~\citep{gardner-etal-2019-making}.

Contemporary cognitive science can be seen as adopting a more holistic approach; integrating symbolic, embodied and distributional accounts~\citep{Lupyan2019}, but also focusing on the crucial \textit{ecological} component~\citep{gibson1979ecological,Hasson2020}: cognition emerges from brain-body-environment interaction. Systematic regularities in the interactions play a key role in inducing systematic linguistic~\citep{narayanan1997knowledge} and knowledge~\citep{Davis2020} representations. These interactional regularities differ in fundamental ways from statistical regularities available to current general NLU methods~\citep{Hasson2020}, for example including perceptual, spatiotemporal and causal dynamics~\citep{Rodd2020,Davis2020}.

Situated (grounded) approaches~\citep{Mikolov2015ART,Liang2016SP} focus on mapping language to executable forms, and highlight the importance of external environments~\citep{McClelland2019}; \citet{Hill2020Environmental} show the emergence of systemic generalization to be contingent on careful task/environment design, rather than specific architectural engineering alone. However, while such environments clearly play an important role in building NLU systems, they are (1) relatively narrow and fixed in terms of semantics (2) costly to create, especially multi-modal environments. 

Here we propose an approach to address this limitation and extend grounded language approaches towards more general domains, by harnessing the power of language to also create and shape environments, rather than just to induce literal execution within them. In this important, yet relatively unexplored role, language helps structure semantic knowledge and serves as a proxy for expensive embodied experience~\citep{Lupyan2016}. To efficiently accomplish this remarkable feat, humans use the language of \textit{affordances}~\citep{gibson1979ecological,Glenberg2008} to construct ``mental worlds''; shaping interactions by specifying what can be done in various situations, from concrete to abstract. We propose that NLU systems should learn to understand (parse) and use such language (e.g., ``This bag can hold up to 20kg before bursting'', see \S\ref{sec:challenges}), which we suggest has a natural programmatic equivalent in the behavioral programming paradigm, such as interactive fiction languages.

In summary, we make the following more concrete contributions and proposals:
\begin{compactitem}
    \item Ecological Semantics: Outline for a theoretical and practical approach to a semantic parsing framework for creation as well as interaction with environments through language. Design considerations are informed by contemporary cognitive science, AI/NLU research and programming language theory (PLT).
    \item We propose methods to inject rich, \textit{actionable} external knowledge into the framework at scale, building upon data mining and automated knowledge base construction (AKBC) research.
    \item We make available\footnote{\url{https://eco-sem.github.io/}} simple interactive demonstrations as working examples showing how such methods can be applied towards  open challenges such as common-sense %
    and causal reasoning.

\end{compactitem}{}

\section{Motivating Challenges: Incorporating World Knowledge}
\label{sec:challenges}

\begin{figure}[t!]
\centering
\includegraphics[width=1\textwidth]{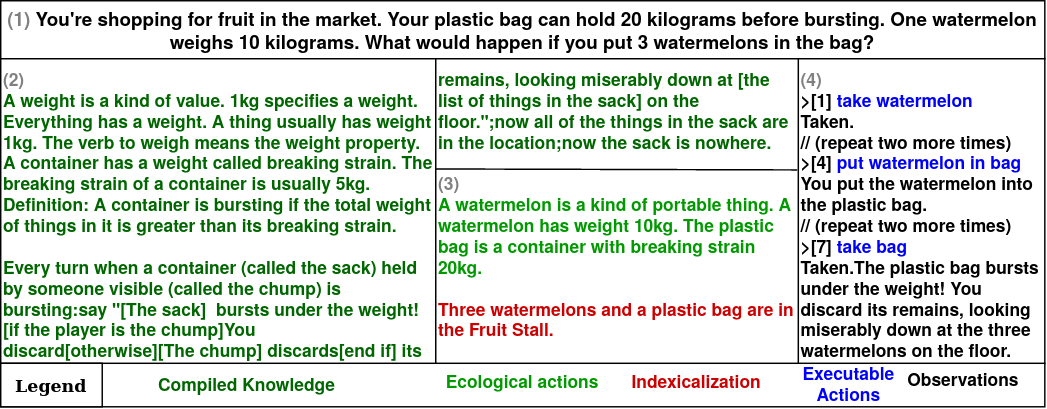}
\caption{\label{fig:watermelons} Inform7 ecological semantic parsing example for \S\ref{sec:challenges} challenge. (1) Input prompt (2) Pre-existing, compiled knowledge (3) Situation knowledge: simulation configuration and indexicalization of referent objects (4) Run simulation to answer ``what-if'' question. 
}
\end{figure}

\xhdr{Explicitly Provided Knowledge} Consider the example in figure \ref{fig:watermelons}, describing an everyday situation of shopping for fruit in a market. Completely trivial for humans, current NLU methods find such ``what-if'' questions highly challenging, even though the relevant affordances are made explicit in the text. A textual entailment model judges it very likely that ``The bag bursts.'' for $X\in\left\{ \text{no,one,two,three}\right\}$\footnote{\url{https://demo.allennlp.org/textual-entailment/}}.

\xhdr{Assumed World Knowledge} In this common, yet more difficult setting, the relevant knowledge is implicitly assumed.  Consider a prompt like ``He put on a white t-shirt and blue jeans. Next, he wore \_''. A completion produced by GPT-2~\citep{radford2019language} is ``a gray cowboy hat, black cargo pants, and white shoes. He also had a black baseball cap pulled low over his eyes''\footnote{\url{https://talktotransformer.com/}}.

Common-sense knowledge graphs are likely to be insufficient for such problems; as shown in ~\citet{forbes2019neural}, ``neural language representations still only learn associations that are explicitly written down'', even after being explicitly trained on a knowledge graph of objects and affordances. As suggested by the work, mental simulations are crucial to common-sense in humans~\citep{Battaglia2013}, allowing the dynamic, affordance-guided construction of relevant representations at run-time as needed, rather than wasting valuable space in memorizing large, ever-incomplete relation graphs.

Importantly, the first problem should be simpler than the second: the required background knowledge is made available in the text. It would be highly desirable to be able to act upon such information. Recent work has begun to explore such capabilities~\citep{Zhong2020RTFM}, but current methods are largely limited in this respect~\citep{Luketina2011}. In the following section, we propose a general problem formulation for incorporating affordances, building upon cognitive linguistics theory.

\section{Ecological Semantics}
\label{sec:eco_sem}
Mental simulations and affordances feature centrally in contemporary cognitive linguistics research. According to one such theory, the Indexical Hypothesis~\citep{Glenberg2008}, language comprehension involves three key processes: (1) indexing objects, (2) deriving their affordances, and (3) meshing them together into a coherent (action-based) simulation as directed by grammatical cues. Importantly, affordances generally cannot be derived directly from words, but rather rely on context and pre-existing object representations. 

\subsection{Computational Formulation}
The Indexical Hypothesis (IH) can be formulated naturally within the model-based framework used in general AI mental simulation research~\citep{Hamrick2019}. At the core of such frameworks is the partially observable Markov Decision Process (POMDP)~\citep{kaelbling1998planning}, which governs the relations between states ($s$), actions ($a$), observations ($o$) and rewards ($r$). Specifically, we focus on the recognition\footnote{Commonly denoted $O^{-1}$, we denote here by $I$ for Indexicalization.} $I:O\to S$, transition  $T:S\times A\to S$ and policy $\pi:S\to A$ functions.

Pre-existing knowledge regarding the environment (objects and their affordances) can be seen to be primarily represented by $T$, with the emulator model being the neural correlate~\citep{Grush2004,Glenberg2008}. In the POMDP formulation, for a linguistic input (or observation) $\utt$, IH can be formulated as (1) compose an initial state representation $\indutt=s_0$ of objects (we assume the simple case where all objects are mentioned in $\utt$) (2) derive affordances, or the set of actions that can be taken in the current situation (3) enact mental simulation by applying $T$ with chosen action. Typically $\utt$ is composed of multiple utterances $\left(\bar{x}_{1},...,\bar{x}_{N}\right)$ and so the simulation may be composed of multiple actions $\rva=\left(a_{0},...,a_{L-1}\right)$. Slightly abusing notation, we can denote the full execution $T\left(s_{0},\rva\right)$ which yields a result state $s_L$. IH can be seen as corresponding to the standard setting in executable semantic parsing/grounded NLU works~\citep{Long2016}:

\xhdr{Executable Semantic Parsing (Ex-SP)} Given a linguistic input $\utt$ and target intent (goal state) $g^{*}$, output action sequence $\rva$ such that $T\left(\indutt,\rva\right)=g^{*}$. Most grounded/executable approaches assume a fixed, programmatic, domain-specific $T$ (navigation environments, SQL engine, etc.) and focus on learning a policy mapping from $\utt$ to $\rva$.

Our proposal thus focuses on ``pushing the envelope'' of $T$ to allow grounded understanding of more general language. IH discusses the comprehension process in cases where the relevant object and affordance information already exists. But how do we learn such representations in the first place? Embodied experience is one way, but a costly and slow one, so here we focus on the role of language in shaping affordance knowledge, specifically modal language, like ``All watermelons are portable.'' Such language can more naturally be seen as modifying\footnote{This is a delicate point- we currently assume the modification is valid, and leave a more thorough discussion of the rules governing what is possible to future work.} the emulator $T$. Therefore, we propose extending the representation of $T$ to allow it to change in time, $T^{\left(t\right)}$, modified by special \textit{eco-actions} $\cact$. These do not change the current state, but rather only the executor (example in fig. \ref{fig:watermelons}). We denote regular executed actions as $\regact$, and a \textit{scenario} (containing possibly both $\regact,\cact$ actions) as $\cactset$. The full execution is then $\intexc\left(s_{0},\cactset \right)$, which denotes applying $T^{\left(t\right)}$ at each timestep.

\xhdr{Ecological Semantic Parsing (Ec-SP)} Given a linguistic input $\utt$ and target intent (goal state) $g^{*}$, output action sequence $\cactset$ such that $\intexc\left(\indutt,\cactset\right)=g^{*}$.

Figure \ref{fig:watermelons} shows how Ec-SP can be utilized towards addressing the challenge problem from \S\ref{sec:challenges}, which is not handled by current Ex-SP methods, as the input language is out-of-domain (so a specific executor would need to be created). The implementation uses Inform7~\citep{Nelson2005}, an interactive fiction (IF) language (see \S\ref{sec:affordance_lang}). Interactive versions of the examples from \S\ref{sec:challenges} are available online.

We distinguish between compiled knowledge vs. situation knowledge: the former refers to existing knowledge encompassed by the emulator (analogous to code libraries that just need to be imported), the latter is new knowledge defined online using eco-acts (analogous to writing a new program). Clearly, a core issue to be managed is the scalable and incremental growth of the emulator: as in regular programming, recurring ecological information (such as watermelons being portable) should become part of the library, rather than having to be re-defined anew in every situation.

\section{Affordable Affordances: Towards Implementation}
\label{sec:affordance_lang}

\textbf{Programmatic emulation of environments} requires an appropriate programming formalism with which environments can be flexibly constructed and configured\footnote{This preliminary approach is naturally biased towards literal language, which is easier to simulate than more abstract language. While a detailed analysis is out of scope, we note that literal language is seen to lay the neural foundations for abstract language understanding~\citep{lakoff1980metaphorical,Davis2020}}. Our focus here is on purely text based construction, from considerations of scale, to remain broadly applicable to general NLU; multi-modal integration is an interesting future direction.
We suggest that a natural paradigm for such a purpose is Behavioral Programming~\citep{Harel2012}, which can also be seen to include certain IF languages, like Inform7~\citep{Nelson2005}. These languages are designed to be reminiscent of natural language, and express semantics in terms of interactional affordances (indeed often using modal verbs like \textit{can}, \textit{mustn't})~\citep{Harel2012}. Current frameworks for creating custom IF training environments~\citep{cote18textworld,tamari-etal-2019-playing} require extensive re-configuration for new domains, and games must be pre-compiled rather than generated dynamically from textual inputs. Most current IF works focus on solving existing games~\citep{Jain2019} or game construction for human entertainment~\citep{ammanabrolu20world}.

\xhdr{Learning emulators at large-scale} This task is closely related to the grand AI challenge of common-sense learning. In humans, common-sense is hard-coded through rich experience~\citep{Hasson2020}; it is reasonable to expect that approximating human emulators will require extensive hard-coding as well. In rendering this task tractable, We join \citet{Kordjamshidi2018} in advocating a tighter loop between learning and programming to represent knowledge: AI should be extensively utilized in hard-coding its own common-sense. Whereas earlier approaches typically consisted of non-executable, relational knowledge graphs (KGs)~\citep{speer2017conceptnet}, in our case knowledge can be represented by code, executable in interactive simulations. KGs will likely be useful for populating an initial ``seed emulator'', as will AKBC methods for learning object~\citep{Elazar2019} and action~\citep{Forbes2017} properties at scale. In \citet{pustejovsky-krishnaswamy-2018-every}, multimodal simulations are used to evaluate automatic affordance extraction. In \citet{Balint2017}, game designers (for human games) utilized NLU methods for learning object affordances. Finally, as symbolic knowledge is by nature incomplete, it will need to be superseded by geometric, multi-modal knowledge representations~\citep{gardenfors2014geometry,pezeshkpour-etal-2018-embedding}.

By affording NLU systems with the ability to programmatically emulate environments in the context of both online discourse comprehension, as well as large-scale, offline common-sense knowledge mining, we hope to advance research efforts towards grounded, general NLU.

\subsubsection*{Acknowledgments}
We thank the Hyadata Lab at HUJI, and Yoav Goldberg, Ido Dagan, and the audience of the BIU-NLP seminar for interesting discussion and thoughtful remarks. The last author is funded by an ERC grant on {\em Natural Language Programming}, grant number 677352, for which we are grateful. This work was also supported by the European Research Council (ERC) under the European Union's Horizon 2020 research and innovation programme (grant no. 852686, SIAM) and NSF-BSF grant no. 2017741 (Shahaf).

\bibliography{my_bib}
\bibliographystyle{iclr2020_conference}

\end{document}